\documentclass{llncs}
\usepackage{graphicx, caption}
\usepackage{subfig}
\usepackage[square,numbers]{natbib}
\usepackage{multirow}
\usepackage{makecell}
\usepackage{booktabs}
\usepackage{hyperref}

\begin{document}
\title{Using Machine Learning To Fuse Verbal Autopsy Narratives And Binary Features In The Analysis Of Deaths From Hyperglycaemia}
%
%
\author{Thokozile Manaka\inst{1}\orcidID{0000-0001-9910-4480} \and
Terence van Zyl\inst{2}\orcidID{0000-0003-4281-630X} \and
Deepak Kar\inst{3}\orcidID{0000-0002-4238-9822} \and
Alisha Wade\inst{4}\orcidID{0000-0002-1158-2523}
}
\authorrunning{T. Manaka et al.}
%
\institute{School of Computer Science and Applied Mathematics, University of The Witwatersrand, Johannesburg\\ 
\email{thokozilemanaka@wits.ac.za} \and
Institute for Intelligent Systems, University of Johannesburg, Johannesburg\\
\email{tvanzyl@uj.ac.za} \and
School of Physics, University of The Witwatersrand, Johannesburg\\
\email{deepak.kar@wits.ac.za} \and
MRC/Wits Rural Public Health and Health Transitions Research Unit, School of Public Health, University of the Witwatersrand, Johannesburg\\
\email{Alisha.Wade@wits.ac.za}}

\maketitle              
\begin{abstract}
Lower-and-middle income countries are faced with challenges arising from a lack of data on cause of death (COD), which can limit decisions on population health and disease management. A verbal autopsy (VA) can provide information about a COD in areas without robust death registration systems. A VA consists of structured data, combining numeric and binary features, and unstructured data as part of an open-ended narrative text. This study assesses the performance of various machine learning approaches when analyzing both the structured and unstructured components of the VA report. The algorithms were trained and tested via cross-validation in the three settings of binary features, text features and a combination of binary and text features derived from VA reports from rural South Africa. The results obtained indicate narrative text features contain valuable information for determining COD and that a combination of binary and text features improves the automated COD classification task.
\end{abstract}
\section{Introduction}
In 2014, the International Diabetes Federation (IDF) \citep{b1} estimated 175 million undiagnosed cases of diabetes. This lack of diagnosis is compounded by the absence of statistics on diabetes-related deaths in many lower-and-middle-income countries (LMIC). The mortality data that is available is frequently derived from hospital records, and it possibly understates mortality from fatalities that occurring outside of hospital \cite{Blackstock}. People with diabetes may develop chronic complications such as neuropathy, nephropathy and retinopathy, uncontrolled hyperglycaemia, and can also develop acute potentially fatal complications such as diabetic ketoacidosis (DKA) and hyperosmolar hyperglycaemic syndrome (HHS) \cite{Gerich, Umpierrez}.

A verbal autopsy (VA) is a technique, endorsed by the World Health Organization (WHO), to determine a likely cause of death (COD) in countries without robust death registration systems \cite{b4}. It is a record of an interview between a non-clinician field worker and a care taker of the deceased about events around an uncertified death. A VA report consists of two kinds of standardized questions: the closed-ended questions where the interviewee responds to a 'yes' or 'no' question and the open-ended section where the interviewee narrates the events around the period of death \cite{b7}. 
Like much electronic data, an analysis of VA reports may benefit from the application of machine learning techniques and while some studies suggest the narrative section of the VA report is unnecessary and of limited use due to its high dimensionality and sparsity, we hypothesized that this narrative text would improve the ability of verbal autopsy reports to predict a cause of death. We compared the binary features, text features and a combination of binary and text features in VA reports from Agincourt, South Africa, for their accuracy in classifying COD from uncontrolled hyperglycaemia. The machine learning techniques of logistic regression, random forest, XGBoost and neural networks were employed to automate COD classification in the three feature settings. We relied on data labeled by physician experts \cite{Blackstock} as our gold standard for determining whether a death had been caused by uncontrolled hyperglycaemia (positive; coded as 1) or not (negative; coded as 0).

\section{Background}
Two-thirds of the 60 million annual deaths in lower-and-middle-income countries occur outside of health facilities, often in countries with weak death registration systems and therefore do not have a recorded cause \cite{b3}. Information about cause of death (COD) is, however, vital for public health researchers and policy makers and verbal autopsy technology was developed to address this need.

Verbal autopsy, a tool in which the caretaker of the decedent is interviewed about their health preceding their death, can be formulated as a semi-supervised learning problem where binary or continuous questionnaire responses form features of the disease and the label of the disease is the categorical response \cite{b22}. VAs have proved to be highly specific ($98\%$) for various types of CODs \cite{b8}.

Several machine learning approaches have been utilized in a number of research, including Nave Bayes, neural networks, support vector machines, random forests, InterVA4, InsilicoVA, AdaBoost, XGBoost, and Tariff \cite{b11, Fottrell, Fottrell2, Danso2} and have demonstrated the ability to automate the coding of COD from binary features of VA reports. While some studies suggest the narrative section of the VA report is unnecessary and of limited use in COD determination \cite{King}, others argue that the narrative section is more convenient as it does not require a speciﬁc questionnaire format and takes less time to collect compared to the closed-ended questions \cite{b20}. Information in free-text narrative, such as the history of the disease symptoms and treatment, may often be essential to making a correct diagnosis.

\citet{Boag} evaluated common representations of clinical text in a variety of clinical tasks and demonstrated that the best method of representing narratives from clinical text for prediction remains debatable. This is due to the fact that word embeddings do not apprehend medical knowledge but describe the meaning of a word based on its context. \citet{Grnarova} also showed that no text representation performs better than all others and that simple representations outscore advanced representations on tasks like age prediction, hospital admission while complex models perform best on diagnosis and length of stay tasks.

Clinical text representation models, ClinicalBERT \cite{Huang}, MeDAL \cite{Wen}, Publicly Available Clinical BERT Embeddings \cite{Alsentzer}, BioBERT \cite{Lee}, BioELMo \cite{Jin} and SciBERT \cite{Beltagy} have been created to improve natural language processing tasks of predicting mortality and hospital readmission from clinical notes. Datasets on which these state-of-the-art narrative text representations are trained include a collection of clinical and electronic health records from hospitals, MIMIC-II (Medical Information Mart for Intensive Care) \cite{Johnson}, MIMIC-III, PubMed abstracts and PMC full text articles and Wikipedia. The training tasks are disease diagnosis, hospital readmission and mortality prediction.

Other work has focused on natural language processing in VA narratives and numerous methods have been used to classify CoD from VA narratives including term frequency and TF-IDF (term frequency–inverse document frequency) \cite{Danso}, the Tariff method \cite{b16}, the neural network classifier \cite{b20} and random forest \cite{AbrahamRF}. \citet{Danso2} demonstrated improved performances of classification algorithms when using linguistic features of part-of-speech tags, noun phrases and word pairs for COD diagnosis from VA reports in addition to frequency-based features. \citet{b21} showed that using models with key phrases as additional features outperforms topic models which depend on features present in the corpus when using a multi-task learning model to learn COD categories. Addition of character information appears to improve model classiﬁcation for smaller datasets (500 to 1000 records) calling for character-based convolutional neural networks (CNNs) and is a promising technique for automated VA coding \cite{b18}.

While the bag of words model excels at tasks that require it to predict categories that are directly represented by single words in their notes, such as the frequency with which the words 'diabetes' and 'sugar' might appear in a VA report. This may make it a good model to predict diabetes as the cause of death for a VA case, however, this approach ignores the order of words inside a document and so fails to reveal all of the underlying conditions behind a patient's death.

Doc2vec has successfully been used to classify documents in various fields including health care \cite{doc2vec, Boag}. Its variants, the continuous distributed bag of words paragraph vector (PV-DBOW) and the distributed memory paragraph vector (DM-PV) work on the idea that the element values of a word are influenced by the values of other words in its vicinity, and this idea is embodied as a neural network structure as with word2vec. On sentiment analysis and text classification tasks, the model surpass the bag of words model and advanced models based on recursive and recurrent neural networks, which is why it was chosen for this study. While both the binary and VA text representation techniques for COD automation outlined above used raw VA reports, we used the traditional approach of physician evaluation of VA forms to incorporate clinical knowledge of diabetes and uncontrolled hyperglycemia into our embeddings. 

The outcomes of this work will aid the public health sector in areas lacking adequate death registration systems by improving automation of COD from binary and text features of VA reports for research, surveillance, and diagnosis of diabetes. Improved VA text representations will provide accurate information regarding diabetes as a cause of mortality, an early detection of which will prevent the lethal complication hyperglycemia.

\section{Methodology and Research Instruments}

\subsection{Algorithms}
\begin{description}
    \item[{Natural Language Processing Algorithms}:] Doc2vec \cite{doc2vec} with both its variants, the continuous distributed bag-of-words paragraph vector (PV-DBOW) and the distributed memory paragraph vector (PV-DM).
    \item[{Classification Algorithms}:] Logistic regression, artificial neural network, random forest and XGBoost.
    \item[{Hardware and Software}:] All experiments were run with Python v3.6 using Sci-kit Learn \cite{Scikit} for random forest, XGBoost and logistic regression classifiers and Keras \cite{chollet2015keras} with Theano \cite{Theano} backend for neural networks. Text preprocessing was done with the Natural Language Tool Kit NLTK \cite{NLTK}.
\end{description}

\subsection{Dataset}

The MRC/Wits Rural Public Health and Health Transitions Research Unit (Agincourt) \cite{Kahn}, a population health and demographic surveillance system located in rural South Africa, that supports research into the origins and effects of diseases on social transitions and populations, provided the verbal autopsy (VA) dataset for this study. Initial coding of the data was done by a paediatrician with expertise in diabetes who reviewed VAs to identify features suggestive of diabetes or uncontrolled hyperglycaemia. Cases where the reviewing physician was uncertain were reviewed with colleagues with experience in adult internal medicine, diabetes, and endocrinology until an agreement was reached (Table~\ref{tab:VAfeature}). Positive instances (79) were those in which uncontrolled hyperglycemia was a likely cause of death and negative (8619) cases were those in which it was not. 

\begin {table}[htb!]
\caption{Features of the Agincourt verbal autopsy dataset.} \label{tab:VAfeature} 
\centering
\resizebox{\columnwidth}{!}{%
\begin{tabular}{lll}
 \toprule
 \textbf{Feature} & \textbf{Type} & \textbf{Description} \\ 
 \bottomrule\toprule
 Female      & Binary  & The deceased's gender \\ 
 Tuber       & Binary  & \makecell[lt]{The dead had been diagnosed with tuberculosis by a medical facility} \\
 Diabetes    & Binary  & \makecell[lt]{The dead had been diagnosed with diabetes by a medical facility} \\  
 Men-con     & Binary  & \makecell[lt]{The dead had signs of mental confusion and memory loss} \\
 Cough       & Binary  & The dead had a cough \\ 
 Ch-Cough    & Binary  & The dead had a chronic cough \\ 
 Diarr       & Binary  & Diarrhoea was present in the deceased \\ 
 Exc-Urine   & Binary  & The dead urinated excessively \\ 
 Exc-Drink   & Binary  & The dead drank water excessively \\  
 Age         & Numeric & The dead's age \\  
 Description & Text    & \makecell[lt]{Narration of signs and symptoms of the deceased by relative}\\
 Class       & Binary  & Cause of death classification by diabetes\\ 
 \bottomrule
\end{tabular}
}
\end {table}

We classified binary features as symptoms to which responses were 'yes' or 'no'. These included excessive thirst, urination, mental confusion and the rest listed in (Table~\ref{tab:VAfeature}). The narration describing symptoms and events around the death of the deceased was classified as a text feature. 

'Yes' and 'no' responses were code to '1s' and '0s' as well as 'Female' and 'Male,' were converted to '1s' and '0s'. To leverage clinical knowledge and incorporate domain knowledge of uncontrolled hyperglycaemia into VA embeddings, we used a physician-annotated dataset. The words 'diabetes, 'sugar' and their misspelled forms were removed from the VA text to uncover symptoms that distinguish cases with diabetes from those without. Text processing included removing punctuation, special characters, short words and stop words as well as lower casing the text. Each VA case includes a paragraph of 4 to 6 sentences of variable length of narratives of instances surrounding death. We tokenized the final text and tagged each case to its respective class as its unique ID and fed it as input to the doc2vec model.

Taking from the Word2vec \cite{MikolovA} model, given a sequence of words $w_1, w_2,..., w_T$ the doc2vec model is trained to predict one of the word's vector $w_t$ given the other $T-k$ words' vectors.

The model maximizes the average log probability
\begin{equation}
\frac{1}{T} \sum_{t=k}^{T-k} \log p(w_t| w_{t-k},...,w_{t+k})   
\end{equation}

\noindent and the task of prediction is

\begin{equation}
 p(w_t| w_{t-k},..., w_{t+k}) = \frac{e^{{y}_{w_t}}}{\sum_{i} e^{y_i}}   
\end{equation}

\noindent In addition the model also takes a paragraph vector $p_i$ with $i$ identifying which body The authors then propose cost-sensitive classification, which differs from random oversampling in that it provides a more continuous-valued and consistent method of weighting samples of imbalanced training data; random oversampling, for example, will invariably favor some training instances over otherof text $w_1, w_2,..., w_T$ comes from.

Each paragraph and word inside a paragraph were mapped to a unique vector and the vectors were next concatenated for prediction of the next word forming a continuous distributed vector representations for VA texts. The contexts were drawn from a sliding window of a defined length. Stochastic gradient descent and back propagation were used to train both word and paragraph vectors. The size of embedding dimension and the length of the context window (the number of words around the target word), were the top parameters for training Doc2vec embeddings and were both tuned experimentally. The link to the code is available at \cite{thoko2021}.

\begin {table}[htb!]
\centering
\caption{Random forest performance on cause of death classification with feature vectors dimensions from PV-DBOW, PV-DM and their combinations.} \label{Vectors}
  \begin{tabular}{ll|rrrrr} 
  \toprule
  \textbf{Method} & \textbf{Dims} & \textbf{Recall} & \textbf{Precision} & \textbf{F1-Score} & \textbf{AUC-ROC} & \textbf{Accuracy} \\ 
  \bottomrule\toprule
  \multirow{6}{*}{\textbf{PV-DM}}
  &\textbf{50} & 0.3702 & 0.0601 & 0.0831 & 0.6468 & 0.9186 \\
  &\textbf{100} & 0.1226 & 0.0147 & 0.0261 & 0.5297 & 0.9302 \\
  &\textbf{200} & 0.2226 & 0.0282 & 0.0494 & 0.5853 & 0.9420 \\ 
  &\textbf{300} & 0.1619 & 0.0422 & 0.0667 & 0.5674 & 0.9663 \\
  &\textbf{400} & 0.1952 & 0.0528 & 0.0802 & 0.5846 & 0.9675 \\
  &\textbf{500} & 0.1333 & 0.0384 & 0.0582 & 0.5540 & 0.9678 \\\midrule
    \multirow{6}{*}{\textbf{PV-DBOW}}
  &\textbf{50} & 0.8464 & 0.8166 & 0.8243 & 0.9224 & 0.9971 \\
  &\textbf{100} & 0.7964 & 0.6731 & 0.7219 & 0.8964 & 0.9948 \\ 
  &\textbf{200} & 0.8048 & 0.7606 & 0.7794 & 0.9012 & 0.9962 \\ 
  &\textbf{300} & 0.7964 & 0.8024 & 0.7980 & 0.8974 & 0.9969 \\
  &\textbf{400} & 0.7774 & 0.8322 & 0.7930 & 0.8879 & 0.9969 \\
  &\textbf{500} & 0.8607 & 0.8196 & 0.8320 & 0.9296 & 0.9974 \\\midrule
  \multirow{6}{*}{\makecell{\textbf{PV-DM +}\\\textbf{PV-DBOW}}}
  &\textbf{50} & 0.8298 & \textbf{0.8571} & \textbf{0.8324} & 0.9141 & \textbf{0.9972}\\
  &\textbf{100} & 0.7548 & 0.5726 & 0.6192 & 0.8746 & 0.9926 \\ 
  &\textbf{200} & \textbf{0.8405} & 0.7526 & 0.7879 & \textbf{0.9189} & 0.9960 \\ 
  &\textbf{300} & 0.8298 & 0.6504 & 0.7231 & 0.9129 & 0.9946 \\
  &\textbf{400} & 0.8131 & 0.7738 & 0.7781 & 0.9053 & 0.9962 \\
  &\textbf{500} & 0.7798 & 0.7528 & 0.7578 &  08889 & 0.9963 \\
  \bottomrule
  \end{tabular}
\end {table}

\citet{Le} show that combining the two variations of a paragraph vector, namely the distributed memory paragraph vector (PV-DM) and the distributed bag of words paragraph (PV-DBOW) improves performance. We sought to test these predictions and experimented with a number of dimensions for feature vectors on both models and their combination. From Table~\ref{Vectors} the optimal vector dimension is 50 on each model. We next cross-validated the window size using the validation set and the optimal context window size was 9 words, with the 10th one as the one to be predicted. 

\subsubsection{Preprocessing}
To address the imbalanced dataset, a technique that combines the synthetic minority oversampling technique (SMOTE) \cite{SMOTE} and the Tomek Links (T-Links) \cite{Thai} undersampling was used. 

By first defining the minority class vector and then determining the number of nearest numbers (k), SMOTE synthesizes new examples in the minority class from existing examples along the boundary joining all k-nearest neighbors in the minority class. A line is generated and a synthetic point is placed between the minority data points and any of their neighbors. This process is performed for each minority data point and its k neighbors until the data is balanced.

The Tomek Links (T-Links) undersampling approach identifies all the pairs of data points that are nearest to each other but belong to different classes. Two points $a$ and $b$ are termed Tomek Links points if for $a$ as an instance of class $A$ and $b$ an instance of class $B$, the distance $d(a, b)$ between $a$ and $b$ $d(a, b) < d(a, c)$ or $d(a, b) < d(b, c)$ for any instance $c$.

In the preprocessing of the VA data, we combined Tomek-Link undersampling with SMOTE oversampling. On classification tasks, the combination of SMOTE and undersampling approaches (ENN and Tomek Links) has been shown to be beneficial, with better area under a receiver operating characteristic AUC-ROC \cite{Purnajaya}. A 5-fold cross validation step was included in the training process to evaluate the classifiers' performance. The SMOTETomek approach was used to sample VA data in each fold, and the classifier was trained on the training folds and validated on the remaining folds. The ideal hyperparameters of each classifier were found using a grid search technique. 

The performance of the classifiers on the COD classification task with respect to physician expert diagnosis was evaluated using accuracy which is defined as the number of correct predictions from all predictions made and precision, which is defined as the number of deaths correctly identified as deaths due to uncontrolled hyperglycaemia out of all the of positive predictions made. We also used recall, defined as the number of deaths correctly identified as due to uncontrolled hyperglycaemia for all deaths due to uncontrolled hyperglycaemia. 

Both recall and precision capture pertinent cases. A model's precision tells us which cases to ignore, even though they were predicted by our models as deaths due to uncontrolled hyperglycemia. Recall tells us what fraction of deaths due to uncontrolled hyperglycaemia to focus on even through they were predicted by our models as deaths not due to hyperglycaemia. To account for the balance of recall and precision, we used the F1-score, which is a weighted average of precision and recall. We were interested in how well our classifiers predict negative cases and positive cases across different thresholds for labeling a death as due to uncontrolled hyperglycaemia. This was determined by a receiver operating characteristic (ROC) curve. We also used the area under the ROC curve (AUC-ROC) which measures the overall performance of an algorithm.

To understand the relationships captured by the vector representations in the text setting, we trained the principal components analysis (PCA) algorithm \cite{Mackiewicz} to reduce the dimensions of the feature vectors (100-D to 2-D vector space) by integrating feature variances and visualized the results on a plot. 

We performed an error analysis where two endocrinologists independently reviewed the false positive (188) and false negative (10) cases predicted by the best classifier in the text features setting. Where there was disagreement between the two reviewers, consensus was reached through discussion. To uncover predictors of uncontrolled hyperglycaemia, we examined the positive class and false positive predictions word embeddings to determine if they captured the same relationships between words by visualizing their alignments. The positive class was made of a vocabulary of 849 unique words. To allow for better visualizations, we looked at the top 150 words which included those of words describing symptoms of uncontrolled hypeglycaemia. The false positive cases vocabulary was 283 words, all of which were used.

\section{Results}

\begin {table}[htb!]
\centering
\caption{Performance comparison of logistic regression, random forest, xgboost and neural network classifiers on cause of death classification in the binary, narrative text and combined features settings.} \label{CombinedFeatures} 
\begin{tabular}{ll|rrrr}
 \toprule
 \textbf{Features} & \textbf{Classifier} & \textbf{Recall} & \textbf{Precision} & \textbf{F1-Score}  & \textbf{Accuracy} \\ 
 \bottomrule\toprule
 \multirow{4}{*}{\textbf{Binary}}
        &\textbf{Logistic Regression} & \textbf{0.8471} & 0.2110 & 0.3199 & 0.9636\\
        &\textbf{Random Forest} & 0.8000 & \textbf{0.2772} & \textbf{0.3885} & \textbf{0.9741} \\ 
        &\textbf{XGBoost} & 0.7529 & 0.0921 & 0.1637 & 0.9384 \\ 
        &\textbf{Neural Network} & 0.7882  & 0.1069  & 0.1873 & 0.9449 \\
 \midrule
 \multirow{4}{*}{\textbf{Text}}
        &\textbf{Logistic Regression} & \textbf{0.7529} & 0.2801 & 0.4067 & 0.9824 \\
        &\textbf{Random Forest} & 0.7412 & 0.2197 & 0.3377 & 0.9770 \\
        &\textbf{XGBoost} & 0.6000 & \textbf{0.6437} & \textbf{0.6166} & \textbf{0.9942} \\
        &\textbf{Neural Network} & 0.5294 & 0.4976 & 0.5001 & 0.9914 \\
 \midrule
 \multirow{4}{*}{\textbf{Combined}}
        &\textbf{Logistic Regression} & 0.7294 & 0.3783 & 0.4962 & 0.9882 \\
        &\textbf{Random Forest} & 0.8117 & 0.3281 & 0.4601 & 0.9844 \\ 
        &\textbf{XGBoost} & \textbf{0.8588} & \textbf{0.7170} & \textbf{0.7807} & \textbf{0.9962} \\ 
        &\textbf{Neural Network} & 0.5647 & 0.6107 & 0.5786 & 0.9935 \\
 \bottomrule
\end{tabular}
\end {table}

The results show that the accuracy of the models is higher than precision, recall and F1-score values across all settings. For the binary features setting, Table~\ref{CombinedFeatures} shows that the logistic regression classifier achieved the best performance in terms of recall (0.8471), i.e out of all of the cases for which uncontrolled hyperglycaemia was a likely cause of death, the model was able to classify 85\% as deaths due to uncontrolled hyperglycaemia. The random forest classifier followed next in terms of performance for the metrics of precision and  F1-score. In the text features setting, the logistic regression performed best in terms of recall (0.7529) while the XGBoost classifier performed best in terms of precision, F1-score and accuracy. In the setting of combined binary and text features, the XGBoost classifier performs across all metrics, followed by the random forest classifier in terms of recall and the neural network in terms of precision, F1-score and accuracy.

\begin{figure}[htb!]
    \centering
    \subfloat{{\includegraphics[width=6cm, height=6cm]{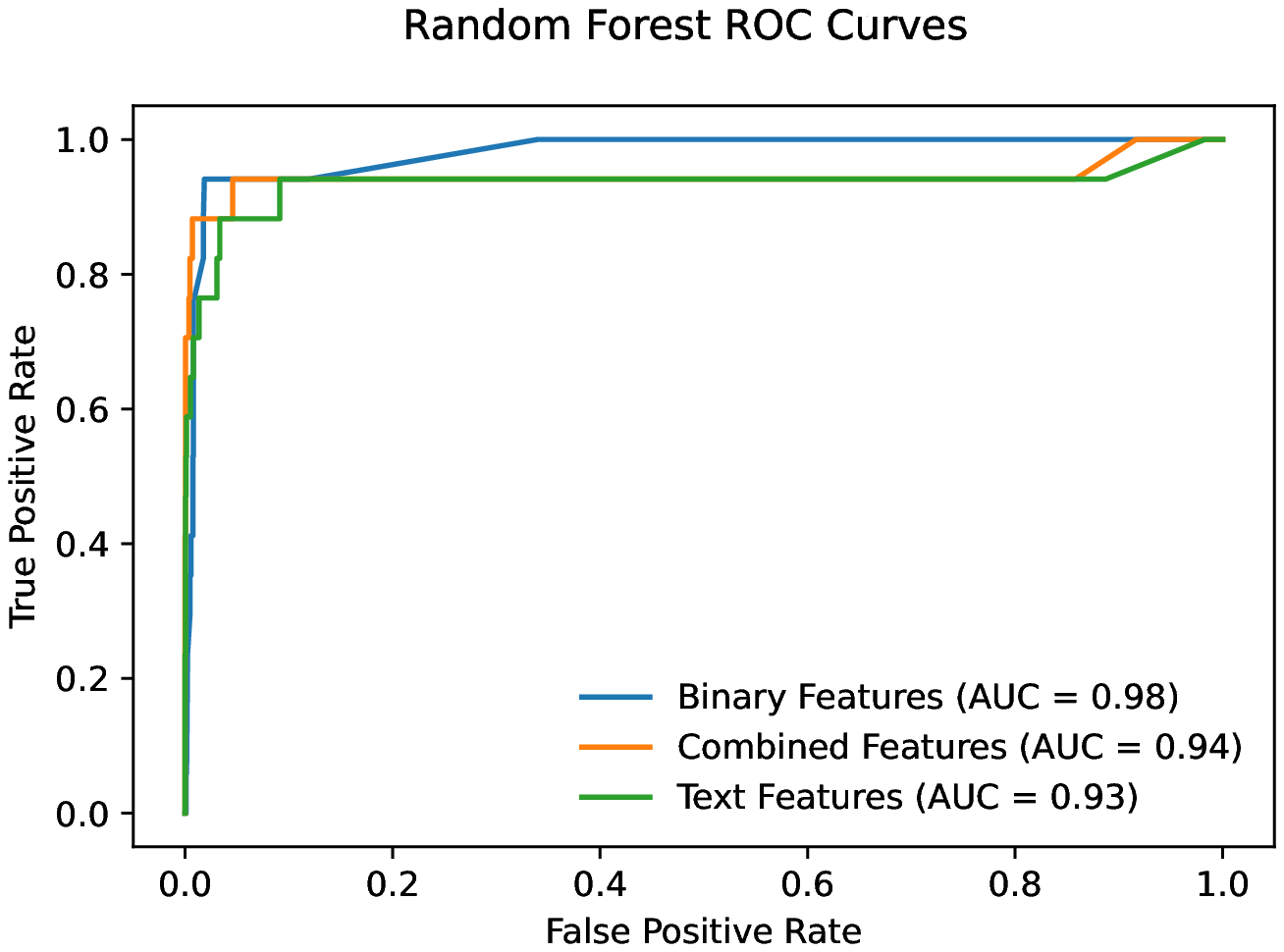}}}
    \subfloat{{\includegraphics[width=6cm, height=6cm]{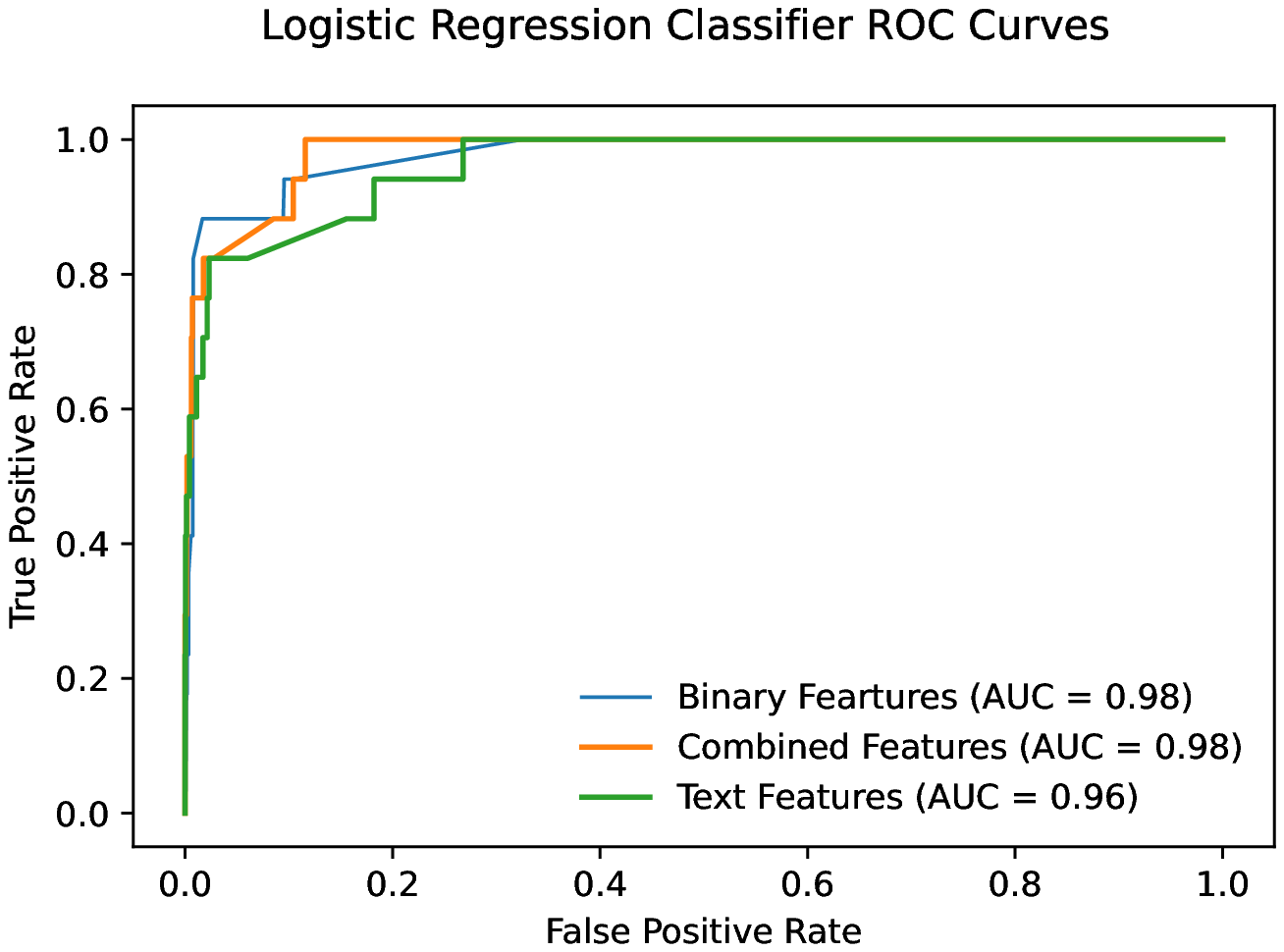}}}
    \caption{ROC curves and AUC-ROC of Random Forest and Logistic Regression classifiers in the binary features, text features and combined binary and text features setting.}
    \label{Figure.1}
\end{figure}

\begin{figure}[htb!]
    \centering
    \subfloat{{\includegraphics[width=6cm, height=6cm]{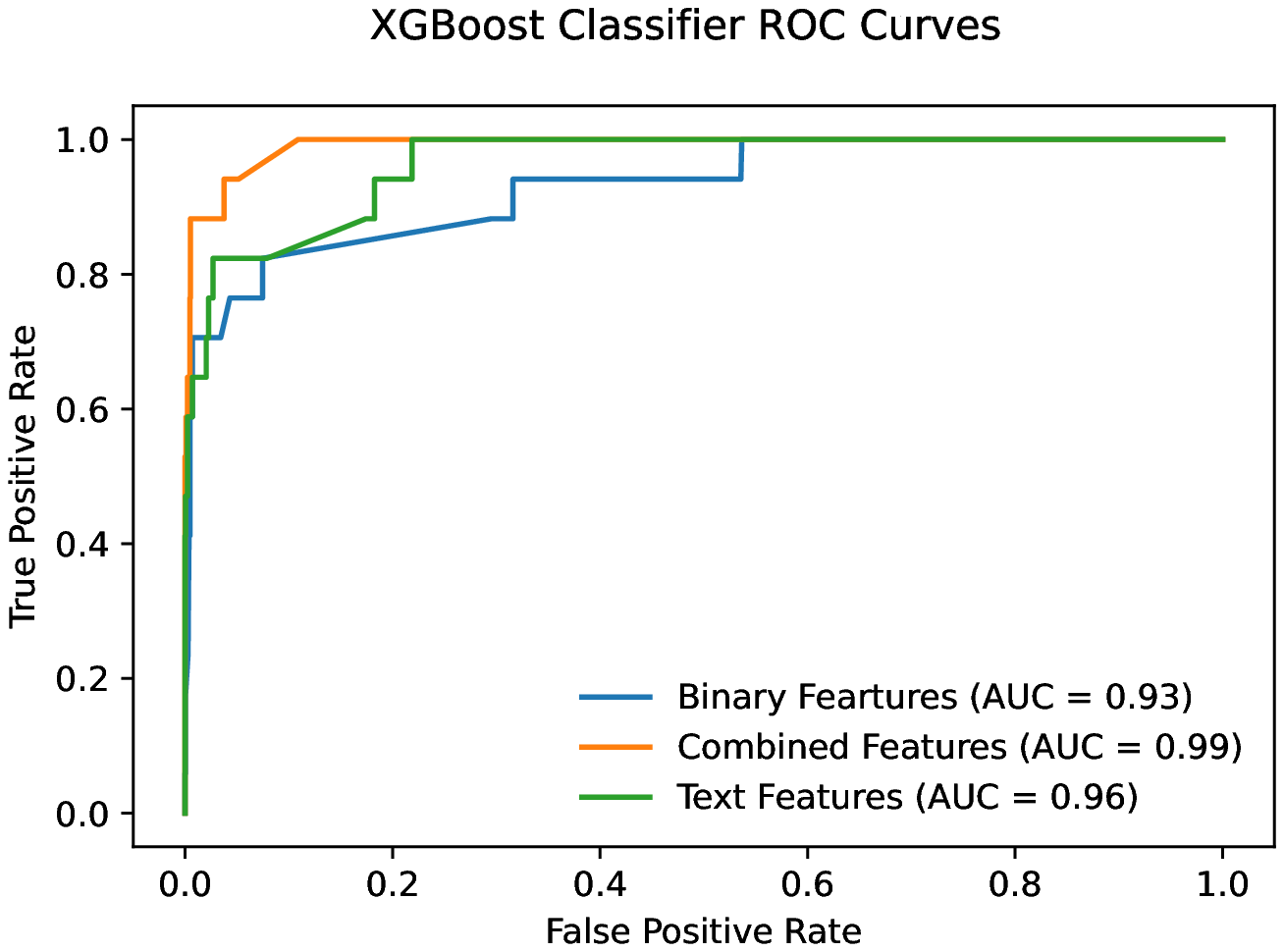}}}
    \subfloat{{\includegraphics[width=6cm, height=6cm]{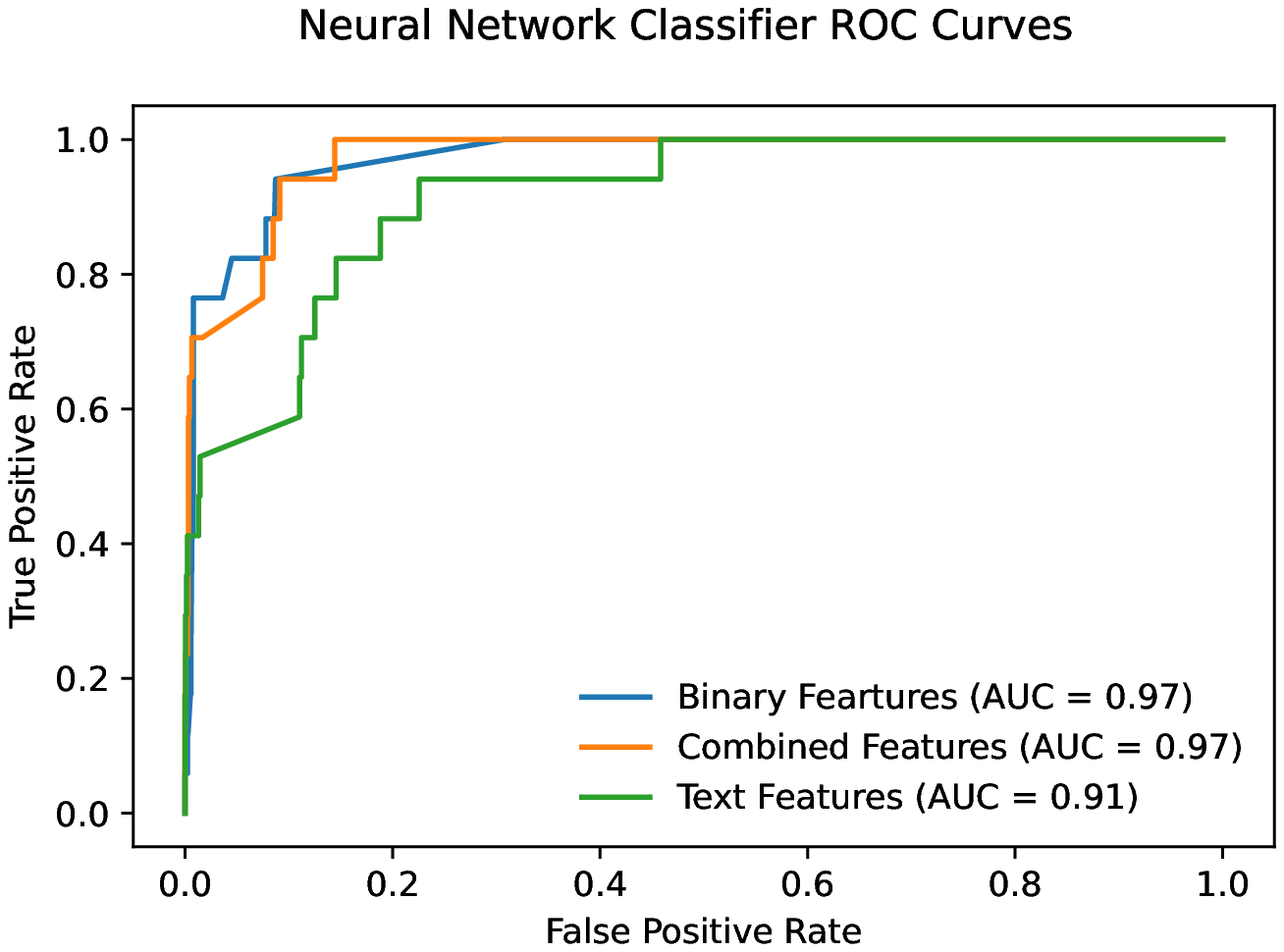}}}
    \caption{ROC curves and AUC-ROC of XGBoost and Neural Network classifiers in the binary features, text features and combined binary and text features setting.}
    \label{Figure.2}
\end{figure}

\begin{figure}[htb!]
  \centering
  \includegraphics[width=\textwidth]{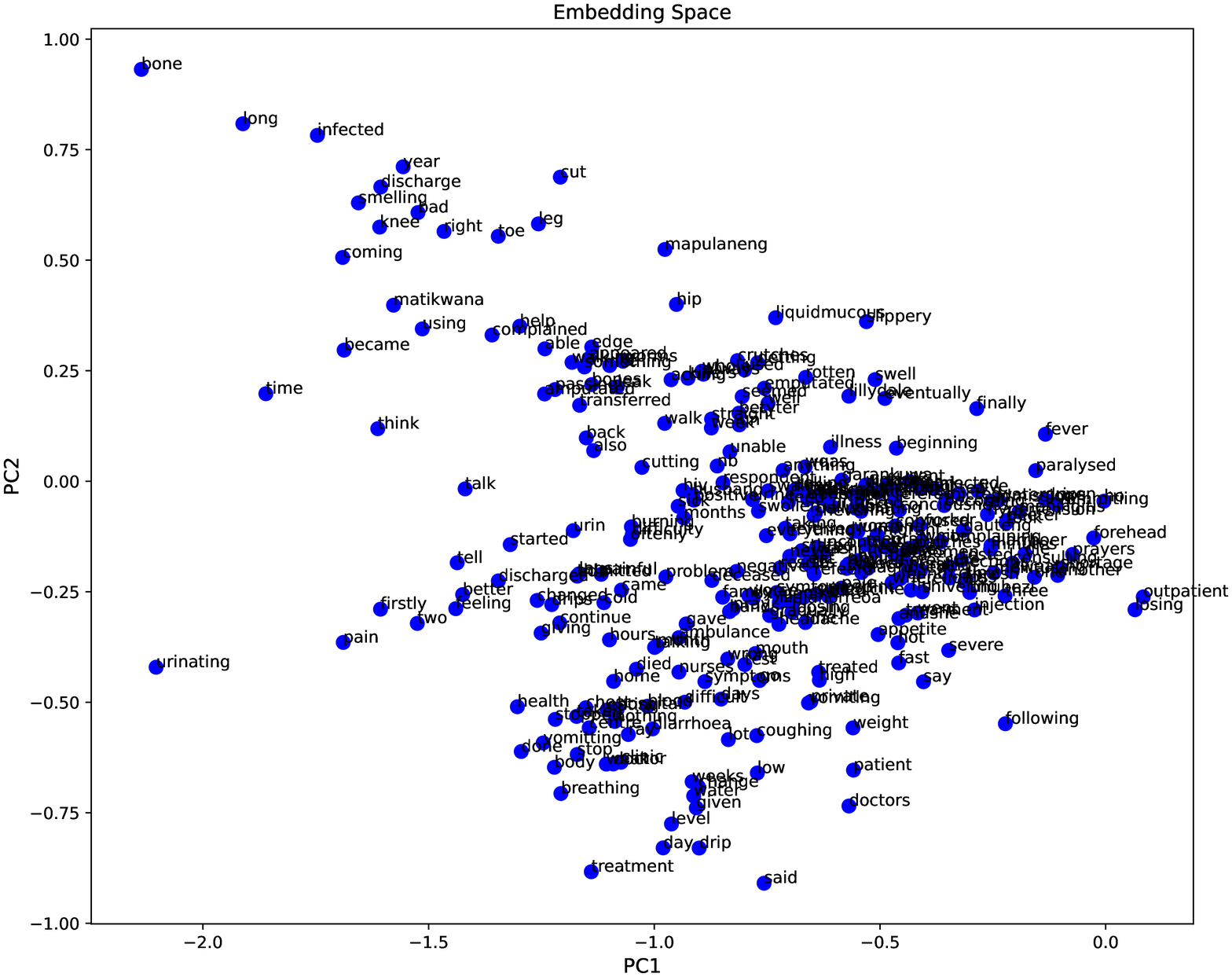}
  \caption{PCA plot: The distribution of keywords indicative of death due to uncontrolled hyperglycaemia.}\label{Figure.3}
\end{figure}

Of interest were high recall scores for all classifiers, but they came at the cost of precision across all the settings except the XGBoost in the text features setting. The best model was then selected by taking the balance between precision and recall (F1-Score) into consideration. In the binary features setting, the best classifier was the random forest while the XGBoost performed best in the text and combined binary and text features settings. 
The receiver operating characteristic (ROC) curve plots of the random forest and logistic regression classifiers across the three settings are given by Fig.~\ref{Figure.1} and those of the XGBoost and Neural Network classifiers by Fig.~\ref{Figure.2}. In the three settings, all the classifier ROC curves climbed toward the top left, meaning the models correctly predict positive and negative cases. We aimed to discover a large number of deaths owing to uncontrolled hyperglycemia from the data, thus we were looking for a threshold with the best sensitivity (recall).
In the binary features setting, the logistic regression and the random forest classifier had the best performance at a threshold that gave a true positive rate (TPR) of around 90\%; i.e where 90\% of all deaths due to uncontrolled hyperglycaemia were being identified and a false positive rate (FPR) of around 30\%. Since more than 90\% of deaths from the VA report were not due to uncontrolled hyperglycaemia, operating at this threshold for both classifiers would be ideal. In the text setting the logistic regression and the XGBoost gave the best area under the ROC curve (AUC-ROC) with the best cut-off point at around (TPR~98\%, FPR~24\%) and (TPR~96\%, FPR~30\%) for the two classifiers. For the combined features setting the same classifiers had the best AUC-ROC scores with ideal threshold values for logistic regression at (TPR~96\%, FPR~18\%) and the XGBoost classifier at (TPR~98\%, FPR~10\%).

The PCA projections to the 2-D vector space are visualized in Fig.~\ref{Figure.3} showing that words with similar part of speech (POS) tags are next to each other (i.e spatially correlated in the embedding space) and that the paragraph2vec model learned words by identifying the neighbouring words.

\subsubsection{Error Analysis}
After the review of the false positive cases (i.e those identified by the best classifier as deaths likely due to uncontrollable hyperglycaemia, but not by the physician-coders) and the false negative cases (cases identified by the best classifier as deaths not likely due to uncontrolled hyperglycaemia, but by the physician-coders as deaths due to uncontrolled hyperglycaemia), a consensus was reached that of the hundred and eighty eight false positive cases, six were indeed likely to be due to uncontrolled hyperglycaemia and of the ten false negative cases, three were as identified as deaths not likely due to uncontrolled hyperglycaemia.

Overall, the six correctly classified false positive cases are an 8\% increase in the positive cases and the improved scores from the text features setting over the combined features setting prove the importance of information available in the free text of VA reports, further highlighting the need for both parts of the VA report in identifying a COD. The results further illustrate that the removal of the keywords, 'sugar' and 'diabetes' from the corpus aided in the classifiers learning the underlying symptoms indicative of hyperglycaemia.

\section{Discussion}

The random forest classifier achieved the best performance for the binary features setting and the XGBoost performed best in the text and combined features settings. While both techniques use ensemble learning, the XGBoost algorithm uses gradient boosting to better capture patterns of high-dimensional features in text and combined text and binary data than the random forest classifier, which depends on making prediction by random chance. In comparison to the binary features and text features settings, the combined features setting had higher scores across all metric because the narrative text provides more depth and context on events around death than the binary features alone.

Our results are in agreement with existing literature. \citet{Sakr, Knuiman} did a comparative study of the same techniques trained with the Synthetic Minority Over-Sampling Technique (SMOTETomek technique) to predict mortality from medical reports using binary features. The results of their study indicated that neural networks perform worse than the other techniques. This is further supported by \citet{Knuiman}, who compared artificial neural networks, Naive Bayes, logistic regression, decision trees, and Bayesian network classifiers on myocardial infarction, with logistic regression outperforming the others (AUC-ROC = 0.82).

The additional features in the combined binary and text setting increase the dimensionality of the feature space, which in turn increases the search space and limits the amount of extraction of valuable information from data \cite{Bolon, Pes}. However, \citet{Pes} has shown the effectiveness of ensemble techniques in handling high-dimensional data which was the case with XGBoost and random forest in the text features and combined features settings in our work. \citet{Clermont} demonstrated how the non-parametric techniques of neural networks, random forests and XGBoost performed better than logistic regression in predicting hospital mortality from clinical notes in ICU patients and \citet{Pirracchio} added that this was because the linear and additive relationship of the logistic regression technique cannot constrain the complex processes surrounding a cause of death and its predictors or variables. 

Word embeddings from the positive class and those from the false positive predictions did capture the same relationships between symptoms of uncontrolled hyperglycaemia as depicted by their vector alignments. This explains the false positive predictions by the best classifier predicts false positives which corroborate the 'distributional hypothesis' \cite{McDonald}, which states that words with comparable or related meanings exist in the same context, and embeddings for semantically and syntactically related keywords from the VA corpus were closer together, as expected.

\subsubsection{Limitations of the study}

This study has three main limitations, the first of which is the dataset size and its class distribution. With 79 positive and 8619 negative cases, the VA dataset from Agincourt has the extremely imbalanced and small minority (EISM) problem. Oversampling and undersampling techniques for redressing this imbalance are not as effective when the absolute number of minority examples is this small \cite{David}, and for this reason we believe significant relationships from the data could have been missed. Additionally, the paragraph2vec model is benchmarked on two very large datasets, the IMDB dataset \cite{Maas} (100 000 movie reviews) and Standford Sentiment Treebank dataset \cite{Socher}, vector dimensions of which were 400. Tuning this parameter on our data, we were below the optimal dimension of 100. 

The second limitation was the way in which we extrapolated medical/ disease knowledge concepts into our embeddings. While there is prior research on domain adaptation techniques to health and clinical text for natural language processing, the VA dataset by nature of its collection, transcription and language cannot be described as a clinical document. Due to lack of publicly available VA word embeddings, our analysis was also not validated against any VA data outside of our institution.

The last limitation of this study is interpretability of VA embeddings. It was not straight forward to provide an explanation for our classifier's judgments based on paragraph and word embeddings because like most embeddings, the paragraph2vec encodes but does not distinguish notions of relatedness and similarity \cite{Khattak}. While some symptoms like "excessive drinking" and "excessive urination" are related, they are not similar. For analysis of cause of death by uncontrolled hyperglycaemia, we would be interested in all symptoms of uncontrolled hyperglycaemia and their combinations and not similar words and synonyms.

\section{Conclusion}
The four machine learning techniques may be useful in the automation of cause of death (COD) classification from verbal autopsy (VA) reports. In this study, they could accurately identify COD from uncontrolled hyperglycaemia, a complication of diabetes. Ensemble learning techniques of random forest and XGBoost are highly accurate, sensitive and specific to this task in all settings of binary, text and combined binary and text features. Our results further suggest that the narrative text of the VA report may contain vital information for determining COD and we therefore encourage further studies to incorporate both binary and text features, separately or in combination. In future work, we will use the transfer learning adaptation techniques of feature extraction, variable selection and fine tuning to make use of broader language representations from the general health and English language domain to improve and refine representations of text from VA reports.\\ 

\noindent \textbf{Acknowledgements} We would like to thank Prof. Justine Davies and the Agincourt Health and Socio-demographic Surveillance System (HDSS) for the VA dataset. We also acknowledge the assistance of Dr. Faheem Seedat in reviewing cases for the error analysis. We thank the United Nations’ Organization of Women in Science for the Developing World (OWSD) for supporting and funding this work. AW is funded by the Fogarty International Centre of the National Institutes of Health (grant number K43TW010698). This paper describes the views of the authors and does not necessarily represent the official views of the National Institutes of Health (USA).

\bibliographystyle{unsrtnat}
\renewcommand{\bibname}{\leftline{References}}
\bibliography{sample}
\end{document}